\pdfoutput=1

\documentclass[11pt]{article}

\usepackage{acl}
\usepackage{times}
\usepackage{latexsym}
\usepackage{float} 
\DeclareUnicodeCharacter{2062}{}
\usepackage{calc}
\usepackage{titlesec}
\titlespacing*{\paragraph}{1pt}{1pt}{1em}

\usepackage[T1]{fontenc}
\usepackage[utf8]{inputenc}
\usepackage{microtype}
\usepackage{inconsolata}
\usepackage{graphicx}

\usepackage[font=small]{caption}

\usepackage{float}

\usepackage{amsthm}

\usepackage{amsmath}
\usepackage[ruled,vlined]{algorithm2e}

\usepackage{enumitem}

\usepackage{multirow}
\usepackage{adjustbox}
\usepackage{bbm}
\usepackage{wrapfig,lipsum}
\usepackage{xspace}
\usepackage{booktabs,cellspace}  
\usepackage[normalem]{ulem}
\usepackage{makecell}
\usepackage{amssymb}
\usepackage{pifont}
\usepackage{todonotes}
\usepackage{microtype}

\usepackage{cleveref}

\newcommand{\stitle}[1]{\vspace{1ex} \noindent{\bf #1.}}

\usepackage{etoolbox}
\usepackage[tikz]{bclogo}

\usepackage{subcaption}
\usepackage{adjustbox}

\definecolor{msftBlue}{RGB}{0,164,239}
\definecolor{msftGreen}{RGB}{127,186,0}
\definecolor{msftYello}{RGB}{255,185,0}
\definecolor{msftBlack}{RGB}{0,0,0}

\usepackage{tcolorbox} 
\tcbuselibrary{skins} 
\usepackage[T1]{fontenc}

\tcbset{
    userstyle/.style={
        enhanced,
        colback=white,
        colframe=black,
        colbacktitle=gray!20,
        coltitle=black,
        rounded corners,
        sharp corners=north,
        boxrule=0.5pt,
        drop shadow=black!50!white,
        attach boxed title to top left={
            xshift=-2mm,
            yshift=-2mm
        },
        boxed title style={
            rounded corners,
            size=small,
            colback=gray!20
        }
    },
    replystyleg/.style={
        enhanced,
        colback=green!15,
        colframe=black,
        colbacktitle=green!30,
        coltitle=black,
        boxrule=0.5pt,
        drop shadow=black!50!white,
        rounded corners,
        sharp corners=north,
        attach boxed title to top right={
            xshift=-2mm,
            yshift=-2mm
        },
        boxed title style={
            rounded corners,
            size=small,
            colback=green!40
        }
    },
    replystyler/.style={
        enhanced,
        colback=red!15,
        colframe=black,
        colbacktitle=red!40,
        coltitle=black,
        boxrule=0.5pt,
        drop shadow=black!50!white,
        rounded corners,
        sharp corners=north,
        attach boxed title to top right={
            xshift=-2mm,
            yshift=-2mm
        },
        boxed title style={
            rounded corners,
            size=small,
            colback=red!40
        }
    }
}

\newtcolorbox{userquery}[1][]{
    userstyle,
    title=Prompt,
    #1
}

\begin{document}

\title{Med-VRAgent: A Framework for Medical Visual Reasoning-Enhanced Agents}

\maketitle

\begin{abstract}
Visual Language Models (VLMs) achieve promising results in medical reasoning but struggle with hallucinations, vague descriptions, inconsistent logic and poor localization.  To address this, we propose a agent framework named Medical Visual Reasoning Agent (\textbf{Med-VRAgent}). The approach is based on Visual Guidance and Self-Reward paradigms and Monte Carlo Tree Search (MCTS). By combining the Visual Guidance with tree search, Med-VRAgent improves the medical visual reasoning capabilities of VLMs. We use the trajectories collected by Med-VRAgent as feedback to further improve the performance by fine-tuning the VLMs with the proximal policy optimization (PPO) objective. Experiments on multiple medical VQA benchmarks demonstrate that our method outperforms existing approaches. Our implementation is publicly available \url{https://github.com/KwongFuk/Med-VRAgent}.

\end{abstract}

\section{Introduction}

Visual Language Models (VLMs) enable context-aware medical reasoning and have shown strong performance in tasks like radiology report generation~\citep{hartsock2024vlmreview,tanno2024flamingocxr,li2023gpt4v}. However, they remain prone to hallucinations, where outputs deviate from the visual input—posing risks in clinical settings~\citep{chen2025medhallmark,jin2024hiddenflaws}. This issue is exacerbated by the factual unreliability of underlying large language models (LLMs)~\citep{huang2023hallucinationsurvey,zhu2024haluevalwild,pal2023medhalt}, highlighting the urgent need for effective mitigation strategies~\citep{kim2025medicalhallucination,bai2025hallucinationmultimodallargelanguage}.

\begin{figure}[t]
    \centering
\includegraphics[width=\linewidth]{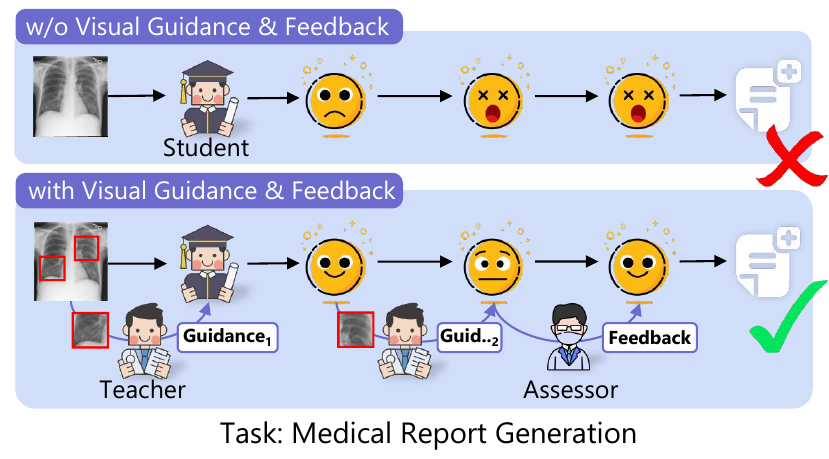}
\caption{Top: A student struggles, feeling confused and making mistakes. Bottom: With guidance, the student overcomes the confusion and successfully completes the task.}
    \label{fig:rg}
\end{figure}

Researchers also have explored several enhancements, the Chain-of-Thought (CoT) has become a popular approach to enhance the logical reasoning capability~\cite{wei2022chain}. Visual prompting—using region-specific cues —have been shown to improve model performance in fields such as radiology and pathology where precise localization is required.~\cite{denner2025visualpromptengineeringvision}. Plan-then-Generate decouple reasoning into structured planning followed by execution~\cite{zhou2022least}. Self-enhancement mechanisms, such as self-reflection, self-correction, and self-critique, and external feedback systems enable models to revise their own reasoning~\cite{madaan2023selfrefine, wang2024qimprovingmultistepreasoning}. Additionally, Retrieval-Augmented Generation (RAG) incorporates external knowledge to support the reasoning process~\cite{lewis2020retrieval}.

While the above approaches are effective, some key challenges remain.  \textbf{(1)} In high-stakes domains like radiology and pathology, VLMs often lack fine-grained image-text alignment, producing overly generic descriptions that miss critical local details, spatial structures, and abnormal patterns. \cite{vishwanath2025medicallargelanguagemodels,liévin2023largelanguagemodelsreason}. \textbf{(2)} Although complex medical prompting strategies have been proposed to address this issue, they are often domain-specific, labor-intensive, and require expert knowledge.~\cite{boiko2023emergent,xia2024carescomprehensivebenchmarktrustworthiness}. \textbf{(3)} Furthermore, current models, even with visual prompting, focus on a single ROI and struggle to integrate overall medical image structure and spatially distributed lesions, limiting performance in cases with high spatial complexity.
~\cite{wang2025exploringinterpretabilityvisualprompt,Huang_2024}.  \textbf{(4)} Current frameworks offer limited feedback, usually evaluating only the final output, making error detection and correction during reasoning difficult. \textbf{(5)} Finally, retrieval enhancement methods often introduce irrelevant or noisy information, potentially distorting clinical reasoning.~\cite{gao2023retrieval,ji2023survey}.

We propose a multimodal agent framework \textbf{Med-VRAgent}, to tackle challenges like error propagation, suboptimal planning, limited feedback, and the fragility of retrieval-based methods. Med-VRAgent consists of three core modules—\textbf{Teacher}, \textbf{Student}, and \textbf{Assessor}—and two key components: a \textbf{Visual Extraction Module}, and a \textbf{Retrieval-Augmented Reflection (RAR)}. The Visual Extraction Module identifies Regions of interest (ROIs) in medical images and uses Visual Token Edit to improve the agent's regional perception. The Teacher provides ROI-specific visual guidance. The Student generate diagnostic outputs with ROI and teacher’s guidance. The Assessor offers fine-grained feedback for iterative refinement.  We use RAR module to enhance factual grounding by incorporating external medical knowledge and introduce  Monte Carlo Tree Search (MCTS) to explore high-quality reasoning paths using an adaptive expansion strategy while better balancing performance and efficiency. Our framework only needs to be trained once for both the teacher and the assessor, which can achieve good generalization ability and save computational resources.         

Results across three benchmarks confirm the superior performance of Med-VRAgent, achieving new state-of-the-art (SOTA) results. It outperforms reasoning baselines (Visual CoT) on GMAI (Table~\ref{tab:results2}), surpasses retrieval-augmented methods on IU-Xray~\cite{DemnerFushman2016} (Table~\ref{tab:results3}), and exceeds advanced fine-tuning strategies like MMedPO on VQA-RAD~\cite{Lau2018VQA} and MIMIC-CXR~\cite{johnson2019mimiccxr} (Table~\ref{tab:results1}). These results highlight the effectiveness of our visual guidance-based medical multimodal agent framework.

In summary, our contributions are as follows:
\begin{itemize}\setlength{\itemsep}{-4pt}
\item We propose a \textbf{Teacher-Student-Evaluator} framework for medical visual reasoning based on \textbf{Visual Guidance} and \textbf{Feedback}.

\item We use \textbf{Visual Extraction} and \textbf{Visual Token Edit} to improve the visual capabilities of multimodal agents.
  
  \item We develop a \textbf{Retrieval-Augmented Reflection} module to further boost agent reasoning via External knowledge.
  
  \item Extensive experiments on multiple medical multimodal benchmarks demonstrate that our framework achieves SOTA performance.
\end{itemize}

\section{Related Work}

\subsection{Foundation Large models}
Large Language Models (LLMs) like GPT-3~\cite{brown2020languagegpt3}, PaLM~\cite{chowdhery2022palm}, and LLaMA~\cite{touvron2023llamav2} have shown strong capabilities in reasoning, generation, and understanding across natural language tasks, excelling in few-shot learning, in-context reasoning, and text generation. These models are central to the development of multi-modal systems. VLMs have demonstrated remarkable generalization across cross-modal tasks such as image captioning, retrieval, and visual question answering (VQA). Early models like CLIP~\cite{radford2021learningtransferablevisualmodels} and Flamingo~\cite{alayrac2022flamingovisuallanguagemodel} use large-scale image-text pairs for contrastive or retrieval-based learning. Recent models like BLIP-2~\cite{li2023blip2bootstrappinglanguageimagepretraining} and MiniGPT-4~\cite{zhu2023minigpt4enhancingvisionlanguageunderstanding} integrate LLMs with visual encoders to enhance reasoning and support open-ended question answering. These advances in Foundation Large Models (FLMs) lay the foundation for tasks that require deep cross-modal understanding.

\subsection{Multi-step Reasoning in FLMs}

\begin{figure*}
    \centering
    \begin{adjustbox}{width=\linewidth}
        \includegraphics[width=\textwidth]{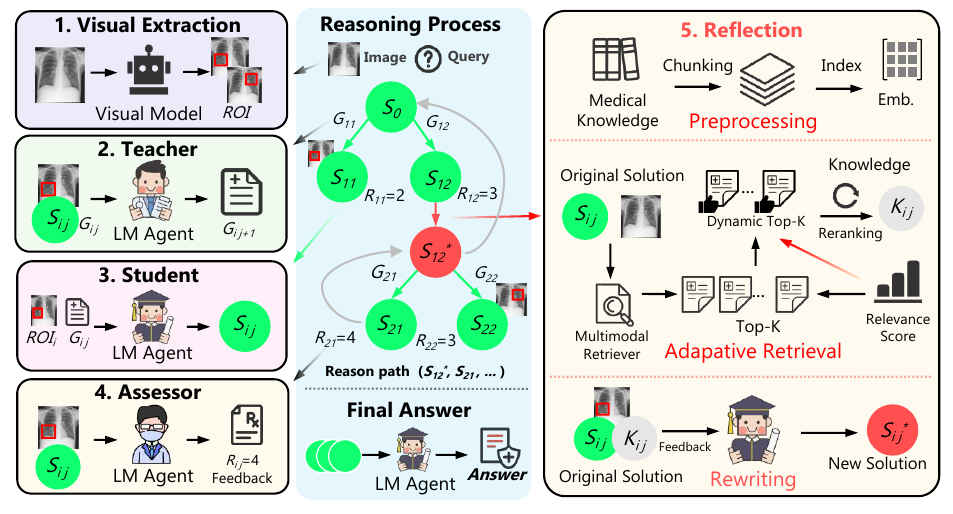}
    \end{adjustbox}
    \caption{Overview of the \textbf{Med-VRAgent} framework. The system uses MCTS to generate solutions \(\mathcal{S}_{ij} \) based on Regions of Interest \(\mathcal{ROI}_{ij}\), visual guides \(\mathcal{G}_{ij}\), rewards \(\mathcal{R}_{ij}\), and external knowledge \(\mathcal{K}_{ij}\).\(\mathcal{S}^*_{ij}\) is the solution after reflection.}
    \label{fig:mcts}
\end{figure*}

Reasoning in Foundation Large Models (FLMs) has advanced with frameworks enhancing multi-step inference and decision-making. CoT~\cite{wei2022chain} enables intermediate reasoning steps, improving performance on complex tasks. ToT~\cite{yao2023treethoughtsdeliberateproblem} explores multiple reasoning paths using tree search strategies, boosting decision-making. The ReAct framework~\cite{yao2022reactsynergizingreasoningaction} combines reasoning with environment interaction, improving tool-augmented tasks. In multimodal reasoning, Visual Chain-of-Thought~\cite{rose2024visualchainthoughtbridging} extends CoT by integrating visual grounding to bridge logical gaps. The Reinforced Ranker-Reader (R$^3$) architecture~\cite{zhang2023active} improves open-domain question answering by combining a ranker and reader with reinforcement learning, optimizing accuracy over retrieved documents.

\subsection{Medical-Specific Reasoning Frameworks}
Medical-Specific Reasoning has advanced with specialized frameworks to enhance LLMs' clinical reasoning. MedAgents~\cite{tang2023medagents} creates a multi-agent system where LLM-based medical experts collaborate on diagnostic tasks, improving zero-shot reasoning. MedReason~\cite{wu2025medreason} aligns LLM reasoning with medical graphs, enhancing decision-making accuracy and interpretability. FineMedLM-o1~\cite{yu2025finemedlm} uses supervised fine-tuning and test-time training on curated dialogues for complex tasks like differential diagnosis. DeepSeek R1~\cite{moell2025deepseek} benchmarks LLM outputs against expert behavior, revealing both advanced reasoning and domain-specific biases. These models highlight the value of tailored frameworks and medical knowledge in improving LLM clinical reasoning.

\section{Methodology}

 To enhance medical visual reasoning, we propose Med-VRAgent, a novel reasoning scheme. It combines a Visual guidance and Reward-Feedback Paradigm in a search algorithm to optimize reasoning paths. 

\subsection{Med-VRAgent Reasoning Process}
Fig~\ref{fig:mcts} illustrates the Med-VRAgent process. We model the agent reasoning process as a tree search, where each node \( \mathcal{S}_{ij} \) represents a state defined as:\noindent%
\vspace{-0.5em}

{\small
\begin{equation}
\mathcal{S}_{ij} = [\mathcal{Q}, \mathcal{I}, \mathcal{G}_{ij}, \mathcal{A}_{ij}, \mathcal{R}_{ij}, \mathcal{F}_{ij}, \mathcal{A}^{*}_{ij}, \mathcal{O}_{ij}, \mathcal{ROI}_{i}]
\end{equation}}
where \( \mathcal{Q} \) is the query, \( \mathcal{I} \) is the medical image, \( \mathcal{G}_{ij} \) is the visual guidance, \( \mathcal{A}_{ij} \) is the current step answer, \( \mathcal{R}_{ij} \) is the reward, \( \mathcal{F}_{ij} \) is feedback, \( \mathcal{A}^{*}_{ij} \) is the answer after reflection, \(\mathcal{O}_{ij}\) represents the observation information, including all ancestor and sibling node guidance and answers, and \( \mathcal{ROI}_{i} \) is the visual ROI. 

Given an image \( \mathcal{I} \) and query \( \mathcal{Q} \), the goal is for Student \( \mathcal{S}_\text{model} \) to generate step-by-step reasoning using ROIs \( \mathcal{ROI}_{i} \) from Vision Extraction \( \mathcal{V}_\text{model} \) and visual guidance \( \mathcal{G}_{ij} \) from Teacher \( \mathcal{T}_{\text{model}}^{\theta} \). Assessor \( \mathcal{A}_{\text{model}}^{\theta} \) evaluates guidance and answers, providing reward \( \mathcal{R}_{ij} \) and feedback \( \mathcal{F}_{ij} \). If answer quality is low, the reflection module uses external knowledge \( \mathcal{K} \) from retriever \( \mathcal{R}_\text{model} \) to refine it. MCTS searches for the optimal reasoning path for answering \( \mathcal{Q} \).

\subsection{Visual Extraction Module}
\paragraph{Visual ROIs Extraction}
We use a lightweight VLM to extract medical entities \( E \) relevant to the question and image. Following MedVP~\cite{zhu2025guidingmedicalvisionlanguagemodels}, we adopt a fine-tuned Grounding DINO~\cite{liu2024groundingdinomarryingdino} as the visual extractor. Grounding DINO is an open-vocabulary detector that localizes entities from image \( I \) and text prompts \( E = \{e_1, e_2, \dots, e_N\} \).%
\vspace{-0.7em}

{\small
\begin{equation}
\text{\(ROI\)} = \{ \text{\(ROI\)}_i \}_{i=1}^{N} = \text{G-DI}(I, E), \quad \text{\(ROI\)}_i = (b_i, s_i, l_i)
\end{equation}
}
\( ROI \) is the set of extracted regions, with each \( ROI_i = (b_i, s_i, l_i) \) representing the bounding box, confidence score, and matched entity label.
\paragraph{Visual Token Edit}
To refine the Agent’s focus on a ROI without retraining, we apply Visual Token Edit (VTE), a single edit to visual tokens in the first (\( K \!\le\! 3 \)) self-attention layers. For each patch embedding \( \mathbf{v}_i \!\in\! \mathbb{R}^d \) and binary ROI mask \( m_i \!\in\! \{0,1\} \), we replace:%
\begin{equation}
\mathbf v_i \;\longrightarrow\;
\mathbf v_i^{\star}= \mathbf v_i + \beta\,m_i\,\mathbf b
\label{eq:VTE_boost}
\end{equation}
where $\mathbf b$ is a fixed direction (e.g., $\mathbf 1$ or $\mathbf v_i$). Because the key and value projections are linear, Eq.\,(\ref{eq:VTE_boost}) increases the $\ell_2$ norm of ROI tokens and thus raises their soft-max attention weights \emph{indirectly}, concentrating information flow on the referenced region while keeping background tokens intact. 

The gain $\beta>0$ is chosen on-the-fly to prevent over- or under-boosting:%
\begin{equation}
\beta = s_i\,
\kappa\,\phi\!\Bigl(\frac{\bar a_{\text{bg}}}{\bar a_{\text{ROI}}}-1\Bigr),
\qquad
\kappa\in[0,1],
\label{eq:VTE_beta}
\end{equation}
where $\bar a_{\text{ROI}}$ and $\bar a_{\text{bg}}$ are the average pre-softmax attention logits of ROI and background patches obtained from a provisional forward pass, and $s_i$ is the detector confidence for the ROI, \(\phi(\cdot)\) is any element-wise activation that is non-negative and monotonically non-decreasing. When the model already attends to the ROI (\(\bar a_{\text{ROI}}\!\ge\!\bar a_{\text{bg}}\)), Eq.\,(\ref{eq:VTE_beta}) yields $\beta=0$, leaving the representation unchanged. Setting $\kappa=0$ disables VTE entirely, making the mechanism safe, computationally negligible, and fully reversible. 

\subsection{Teacher-Student-Assessor  Mechanism}
\stitle{Teacher Agent} 
In natural language tasks, the exponential growth of tag combinations severely limits vanilla MCTS. To improve efficiency, we incorporate a prompt-driven Teacher \( \mathcal{T}_{\text{model}}^{\theta} \) that expands the policy space via heuristics. See the appendix~\ref{fig: system_prompt_1} for prompt. At each node, \( \mathcal{T}_{\text{model}}^{\theta} \) gathers prior guidance–answer pairs \((\mathcal{G}_{1..i}, \mathcal{A}_{1..i})\) and feedback \(\mathcal{F}\), then generates the next-step guidance:
\begin{equation}
\mathcal{G}_{ij+1} = \mathcal{T}_\text{model}(\mathcal{ROI}_i, \mathcal{G}_{i1..j}, \mathcal{A}_{i1..j}, \mathcal{F}_{i})
\end{equation}
\stitle{Student Agent} 
The Student \( \mathcal{S}_\text{model} \) leverages a vision-language backbone to perform step-wise reasoning. At each stage of problem, it receives the Teacher \( \mathcal{T}_{\text{model}}^{\theta} \)-generated guidance \( \mathcal{G}_{ij} \) and the corresponding image  \( \mathcal{ROI}_i \), and produces an intermediate answer \( \mathcal{A}_{ij} \). After search, the best reasoning path selected by MCTS is used to compose the final answer. This process is formally defined as:
\begin{equation}
\mathcal{A}_{ij} = \mathcal{S}_\text{model}(\mathcal{ROI}_i,\, \mathcal{G}_{ij}),
\label{eq:student_model}
\end{equation}
\stitle{Assessor Agent} In the MCTS, it is essential to quantitatively evaluate each reasoning step and provide high-quality feedback to guide the search process. To this end, we adopt a \textit{LLM-as-a-Judge}~\cite{gu2025surveyllmasajudge} approach, we introduce an Assessor model \( \mathcal{A}_{\text{model}}^{\theta} \), implemented using a VLM, and grounded in the \textit{Self-Rewarding} paradigm~\cite{yuan2025selfrewardinglanguagemodels} The Assessor \( \mathcal{A}_{\text{model}}^{\theta} \) employs a 5-point scoring system to evaluate task progress, where the score reflects both the quality and contribution of each intermediate answer. The Assessor \(  \mathcal{A}_{\text{model}}^{\theta}  \) receives the image ROI \( \mathcal{ROI}_i \), the current guidance \( \mathcal{G}_{ij} \), and the student’s answer \( \mathcal{A}_{ij} \). It then produces both a descriptive feedback \( \mathcal{F}_{ij} \) and a quantitative rating \( \mathcal{R}_{ij} \), See Appendix~\ref{fig: system_prompt_2} for prompt. The process is formalized as:
\begin{equation}
\mathcal{F}_{ij},\, \mathcal{R}_{ij} = \mathcal{A}_\text{model}(\mathcal{ROI}_i,\, \mathcal{G}_{ij},\, \mathcal{A}_{ij}).
\end{equation}
\subsection{Retrieval-Augmented Reflection}
The reflection phase is designed to enhance ROI analysis tasks that the Student \( \mathcal{S}_\text{model} \) fails to complete under Teacher \( \mathcal{T}_{\text{model}}^{\theta} \) guidance. We use IU-Xray, MIMIC-CXR, VQA-RAD and other datasets as knowledge sources. The reflection process consists of two stages: 

\stitle{Retrieval Phase}
We adopt the domain-aware retriever from MMed-RAG~\cite{xia2025mmedragversatilemultimodalrag}, which uses ResNet-50~\cite{he2015deepresiduallearningimage} and BioClinicalBERT~\cite{alsentzer2019publiclyavailableclinicalbert} as the image and text encoders, respectively.  During reflection, the retriever takes as input the guidance \( \mathcal{G}_{ij} \), image \( \mathcal{I} \), and answer \( \mathcal{A}_{ij} \). It first retrieves a Top-\( \mathcal{K} \) candidate set \( \mathcal{K}_1 \) from the external knowledge base using FAISS~\cite{johnson2017billion}. A cross-attention-based relevance scoring model cross-encoder/ms-marco-MiniLM-L-6-v2~\cite{reimers2019sentencebert} then refines these candidates into a subset \( \mathcal{K}_2 \), which is finally reranked to produce the final knowledge set \( \mathcal{K}_{ij} \). This multi-stage knowledge retrieval process is formally expressed as:
\vspace{-0.5em}
\begin{equation}
\resizebox{\columnwidth}{!}{$
\mathcal{K}_{ij} = \text{Rerank}\Big(\text{Relevance}\big(\text{RetrieveTop-}K(\mathcal{I}, \mathcal{G}_{ij}, \mathcal{A}_{ij})\big)\Big)
$}
\end{equation}
\vspace{-0.5em}
\stitle{Rewriting Phase}
When reflection is needed, the student \( \mathcal{S}_\text{model} \) receives the original answer \( \mathcal{A}_{ij} \), guidance \( \mathcal{G}_{ij} \), the input \( \mathcal{ROI} \), feedback \( \mathcal{F}_{ij} \), and retrieved knowledge \( \mathcal{K}_{ij} \). It then synthesizes these inputs to produce a refined answer \( \mathcal{A}^*_{ij} \). This rewriting process can be formalized as:
\vspace{-0.5em}
\begin{equation}
\mathcal{A}^*_{ij} = \mathcal{S}_\text{model}(\mathcal{ROI}_{i}, \mathcal{G}_{ij}, \mathcal{A}_{ij}, \mathcal{F}_{ij}, \mathcal{K}_{ij})
\end{equation}

\subsection{ Monte Carlo Tree Search Process}

Monte Carlo Tree Search (MCTS) operates through four main phases—selection, expansion, evaluation, and backpropagation—repeating until satisfactory reasoning results are produced or computational limits are reached. In the \textbf{Selection} phase, the algorithm starts at the root node (initial state \( \mathcal{S}_0 \)) and recursively selects child nodes using the Upper Confidence Bound (UCB) formula, which balances exploration and exploitation:%
\begin{equation}
UCB(s) = \frac{R(s)}{N(s)} + C \cdot \sqrt{ \frac{2 \cdot \ln N(p)}{N(s)} }
\end{equation}
where \( R(s) \) is the reward, \( N(s) \) the visit count of node \( s \), \( N(p) \) the visit count of its parent \( p \), and \( C \) is a constant. The \textbf{Expansion} phase involves selecting an unprocessed ROI along the current path and expanding it by sampling \( \mathcal{N} \) guidance suggestions from the Teacher \( \mathcal{T}_{\text{model}}^{\theta} \). This step incorporates a heuristics mechanism, where feedback from Assessor \( \mathcal{A}_{\text{model}}^{\theta} \) and all observations—including guidance, answer from ancestor and sibling nodes are provided to the Teacher \( \mathcal{T}_{\text{model}}^{\theta} \). In the \textbf{Evaluation} phase, each new child node is assessed using feedback from the Assessor \( \mathcal{A}_{\text{model}}^{\theta} \). Finally, in the \textbf{Backpropagation} phase, the reward \( \mathcal{R}(s') \) is used to update the average reward and visit counts for node \( \mathcal{S'} \) and its ancestors. 

To improve search performance and efficiency in MCTS, we apply some strategy.

\textbf{Early Stopping.} Expansion is terminated when the node score exceeds 4 or when KL divergence and semantic similarity suggest the Student \( \mathcal{S}_\text{model} \) and Teacher \( \mathcal{T}_{\text{model}}^{\theta} \) outputs align with the previous node. This allows the agent to shift to other ROIs.

\textbf{Alpha-Beta Pruning.} During selection and expansion, Alpha (min guaranteed by maximization) and Beta (max guaranteed by minimization) bounds are maintained. Subtrees are pruned when node scores fall outside this range, avoiding unnecessary evaluations.

\textbf{Reflection.} If early stopping is triggered repeatedly or the expansion limit is reached without achieving a score of 4, the reflection module is activated. In this case, the Student \( \mathcal{S}_\text{model} \) retrieves external knowledge to continue reasoning.

\subsection{Training Strategy and Optimization}

To enhance the Teacher \( \mathcal{T}_{\text{model}}^{\theta} \) and Assessor \( \mathcal{A}_{\text{model}}^{\theta} \), we fine-tune both VLMs using proximal policy optimization (PPO) with feedback trajectories collected by Med-VRAgent. PPO optimizes the policy by maximizing expected rewards while constraining updates to avoid performance degradation. The objective is:%

{\small
\begin{equation}
\mathcal{L}_{\text{PPO}}(\theta) = \mathbb{E} \left[ \min \left( r_\theta \hat{A}_t, \, \text{clip}(r_\theta, 1 - \epsilon, 1 + \epsilon) \hat{A}_t \right) \right]
\end{equation}
}
where
\begin{equation}
r_\theta = \frac{\pi_\theta(A_{1..i} | O_{1..i})}{\pi_{\theta_{\text{old}}}(A_{1..i} | O_{1..i})}
\end{equation}
Here, \( A_{1..i} \) and \( O_{1..i} \) denote sampled actions (guidance) and observations, respectively, while \( \hat{A}_t \) is the advantage estimate and \( \epsilon \) is the clipping threshold. We collect trajectories
\begin{equation}
\mathcal{T}_{\text{Med-VRAgent}} = (A_{1..i}, O_{1..i}, R_{1..i})
\end{equation}
from Med-VRAgent to estimate advantages and update the policy parameters \( \theta \). The clipping in \( \mathcal{L}_{\text{PPO}}(\theta) \) ensures conservative, stable updates. 

\section{Experiments}
\subsection{Experimental Datasets}
\begin{table}[htbp]
\centering
\begin{adjustbox}{width=\linewidth}
        \setlength{\tabcolsep}{4pt}
        \tiny
\resizebox{\textwidth}{!}{
\begin{tabular}{lcccll}

\toprule
\textbf{Dataset} & \textbf{Modality} & \textbf{Size}   & \textbf{Task Type} \\
\midrule
IU-Xray       & X-ray         & 590                     & Report Generation \\
MIMIC-CXR     & Chest X-ray   & 500        &  Report Generation \\
VQA-RAD       & X-ray, CT     & 451             & Visual Question Answering \\
GMAI-MMbench & 38 modalities & 4 task      & Visual Question Answering \\
\bottomrule
\end{tabular}
}
\end{adjustbox}
\caption{The medical visual datasets used in this experiment}
\label{tab:med-vl-datasets}
\end{table}

We evaluate Med-VRAgent on various medical visual-linguistic datasets covering report generation and VQA tasks. As shown in Table~\ref{tab:med-vl-datasets}, for report generation, we use the IU-Xray~\cite{DemnerFushman2016} dataset containing 590 test samples  and the MIMIC-CXR~\cite{johnson2019mimiccxr} dataset test with 500 test samples. For VQA, we use VQA-RAD containing test  451 QA pairs based on X-rays and CT images and GMAI-MMbench~\cite{chen2024gmai}  we use 4 clinical tasks. 

For the Med-VQA task, for open questions, we report recall in the Open column. For closed questions, we report precision in the Closed column. For the report generation task, we use BLEU~\cite{papineni-etal-2002-bleu} Score, ROUGE-L~\cite{lin-2004-rouge}, and METEOR as metrics~\cite{banerjee-lavie-2005-meteor}. BLEU score represents the average of BLEU-1/2/3/4.

\subsection{Compared Methods}
 
We evaluate the performance of various methods across different approaches. 

For training methods, we employ the LLaVA-Med~\cite{li2023llavamedtraininglargelanguageandvision} model and assess its performance on the VQA-RAD and MIMIC-CXR datasets. The training approaches compared include SFT, Self-Rewarding~\cite{yuan2025selfrewardinglanguagemodels}, Direct Preference Optimization (DPO)~\cite{rafailov2024directpreferenceoptimizationlanguage}, STLLaVA-Med, and MMedPO~\cite{zhu2024mmedpoaligningmedicalvisionlanguage}. 

For reasoning methods, we use the DeepSeek-VL-7B~\cite{lu2024deepseekvlrealworldvisionlanguageunderstanding} and MiniCPM-V2~\cite{yao2024minicpmvgpt4vlevelmllm} models, evaluating their performance on the GMAI-MMbench. The reasoning approaches compared include CoT~\cite{wei2022chain}, ToT~\cite{yao2023treethoughtsdeliberateproblem}, and Visual CoT~\cite{shao2024visualcotadvancingmultimodal}. 

Finally, for Decoding-based and Retrieval-Augmented methods, we use the LLaVA-Med v1.5 model and evaluate its performance on the IU-Xray dataset. The Decoding-based methods include Greedy Decoding, BeamSearch~\cite{xie2023selfevaluationguidedbeamsearch}, DoLa~\cite{chuang2024doladecodingcontrastinglayers}, OPERA~\cite{huang2024operaalleviatinghallucinationmultimodal}, VCD~\cite{leng2023mitigatingobjecthallucinationslarge}. The RAG approaches compared include MedDr~\cite{he2024meddr}, FactMM-RAG~\cite{sun2025factawaremultimodalretrievalaugmentation}, RULE~\cite{xia2024rulereliablemultimodalrag}, and MMed-RAG~\cite{xia2025mmedragversatilemultimodalrag}. Please see the appendix~\ref{sec:Comparedmethods} for details.

\subsection{Model Implementation }
We applied Med-VRAgent to LLaVA-Med v1.5, DeepSeek-VL-7B, and MiniCPM-V2. To ensure fair comparison, we follow the same experimental settings as prior work, using a decoding temperature of 0.7. We use DeepSeek-VL-7B as the Teacher \( \mathcal{T}_{\text{model}}^{\theta} \) and Assessor \( \mathcal{A}_{\text{model}}^{\theta} \) and perform PPO fine-tuning.

For PPO fine-tuning, We follow the official training scripts and use the "peft" and "trl" Python packages to implement LoRA and PPO. The fine-tuning process is completed within 7–8 hours on 4 Nvidia A6000 GPUs. The "lora\_target\_modules" are set to ["q\_proj", "v\_proj"], with lora\_r set to 16, lora\_alpha set to 32, and lora\_dropout set to 0.05. The micro\_batch\_size is 1, the batch\_size is 8, and num\_epochs is 1. For optimization, we set the learning\_rate to 1.41e-5, the reward baseline to 3.75, and the random seed to 0.

\subsection{Overall Performance}

\begin{table}[t]
    \centering
    \begin{adjustbox}{width=\linewidth}
        \setlength{\tabcolsep}{2pt}
        \tiny
        \begin{tabular}{@{}ll|cc|ccc@{}}
            \toprule
            \multirow{2}{*}{\textbf{}} & \multirow{2}{*}{\textbf{Methods}} & \multicolumn{2}{c|}{\textbf{VQA-RAD}} & \multicolumn{3}{c}{\textbf{MIMIC-CXR}} \\
            & & \textbf{Open} & \textbf{Closed} & \textbf{BLEU} & \textbf{ROUGE-L} & \textbf{METEOR} \\
            \midrule
            \multirow{7}{*}{}
           
            &  LLaVA-Med v1.5   & 29.24 & 63.97 & 10.25 & 9.38 & 7.71 \\
            &  ~SFT              & 31.38 & 64.26 & 12.39 & 10.21 & 8.75 \\
            &  ~Self-Rewarding   & 32.69 & 65.89 & 12.15 & 10.05 & 8.77 \\
            &  ~DPO              & 32.88 & 64.33 & 12.37 & 10.38 & 9.10 \\
            &  ~STLLaVA-Med      & 33.72 & 64.70 & 12.21 & 10.12 & 8.98 \\
            &  ~MMedPO           & 34.03 & 67.64 & 13.28 & 13.22 & \textbf{10.20} \\
            &  ~\textbf{Med-VRAgent (Ours)} & \textbf{35.70} & \textbf{68.72} & \textbf{13.90} & \textbf{13.53} & 9.58 \\
            \bottomrule
        \end{tabular}
    \end{adjustbox}
    \caption{
        Comparison of Med-VRAgent with fine-tuning methods, including SFT, Self-Rewarding, DPO, STLLaVA-Med, and MMedPO, evaluated on VQA-RAD (Open/Closed Accuracy) and MIMIC-CXR (BLEU, ROUGE-L, METEOR) datasets, based on LLaVA-Med v1.5. The best result for each model is bolded.
    }
    \label{tab:results1}

\end{table}

\begin{table}[t]
\centering 

\begin{adjustbox}{width=\linewidth}
        \setlength{\tabcolsep}{2pt}
        \tiny
\begin{tabular}{@{}ll|cccc|c@{}} 

\toprule 
\textbf{} & \textbf{Methods} & \textbf{AR} & \textbf{BVR} & \textbf{B} & \textbf{CR} & \textbf{Average} \\ 
\midrule 
\multirow{5}{*}{} 
& DeepSeek-VL-7B & 38.43 & 47.03 & 42.31 & 37.03 & \textcolor[RGB]{30,60,255}{41.20} \\ 
& ~CoT & 39.24 & 46.60 & 43.26 & 38.18 & \textcolor[RGB]{30,60,255}{41.57} \\ 
& ~ToT & 40.23 & 46.07 & 44.42 & 39.58 & \textcolor[RGB]{30,60,255}{42.08} \\ 
& ~Visual CoT & 41.57 & 46.76 & 44.13 & 41.59 & \textcolor[RGB]{30,60,255}{43.51} \\ 
& \textbf{Med-VRAgent (Ours)} & \textbf{44.81} & \textbf{51.82} & \textbf{47.52} & \textbf{42.79} & \textcolor[RGB]{30,60,255}{\textbf{46.74}} \\ \midrule 

\multirow{5}{*}{} 
& MiniCPM-V2 & 40.74 & 43.01 & 36.46 & 37.57 & \textcolor[RGB]{30,60,255}{39.45} \\ 
& ~CoT & 41.69 & 43.90 & 37.69 & 38.74 & \textcolor[RGB]{30,60,255}{40.51} \\ 
& ~ToT & 42.14 & 44.32 & 38.27 & 39.29 & \textcolor[RGB]{30,60,255}{41.01} \\ 
& ~Visual CoT & 43.20 & 44.70 & 39.12 & 41.28 & \textcolor[RGB]{30,60,255}{42.08} \\ 
& \textbf{Med-VRAgent (Ours)} & \textbf{44.81} & \textbf{47.32} & \textbf{40.18} & \textbf{41.34} & \textcolor[RGB]{30,60,255}{\textbf{43.41}} \\

\bottomrule \end{tabular}

\end{adjustbox}

\caption{
 Comparison of Med-VRAgent with reasoning methods, including CoT, Tree-of-Thought (ToT), and Visual CoT ,evaluated  on the GMAI (Accuracy) dataset, based on DeepSeek-VL-7B and MiniCPM-V2. GMAI include AR (Attribute Recognition), BVR (Blood Vessels Recognition), B (Bone), and CR (Cell Recognition). The best result for each model is bolded, and average values are in blue.}
 \label{tab:results2}

 \end{table}

\begin{table}[t]
    \centering
    \begin{adjustbox}{width=\linewidth}
        \setlength{\tabcolsep}{4pt}
        \tiny
        \begin{tabular}{@{}ll|ccc@{}}
            \toprule
            \textbf{} & \textbf{Methods} & \textbf{BLEU} & \textbf{ROUGE-L} & \textbf{METEOR} \\
            \midrule
            
            \multirow{12}{*}{}
            & LLaVA-Med v1.5       & 9.64  & 12.26 & 8.21  \\
            & ~Greedy          & 11.47 & 15.38 & 12.69 \\
            & ~Beam Search     & 12.10 & 16.21 & 13.17 \\
            & ~DoLa            & 11.79 & 15.82 & 12.72 \\
            & ~OPERA           & 10.66 & 14.70 & 12.01 \\
            & ~VCD             & 10.42 & 14.14 & 11.59 \\
            \cmidrule(lr){2-5}
            & ~MedDr           & 12.37 & 16.45 & 13.50 \\
            & ~FactMM-RAG      & 14.70 & 18.05 & 15.92 \\
            & ~RULE            & 27.53 & 23.16 & 27.99 \\
            & ~MMed-RAG        & 31.38 & 25.59 & 32.43 \\
            \cmidrule(lr){2-5}
            & ~\textbf{Med-VRAgent} & \textbf{33.45} & \textbf{26.81} & \textbf{33.12} \\
            \bottomrule
        \end{tabular}
    \end{adjustbox}
    \caption{
    Comparison of Med-VRAgent with RAG methods, including FactMM-RAG, MMed-RAG etc, on the \textbf IU-Xray (BLEU, ROUGE-L, METEOR) dataset, based on \textbf{LLaVA-Med v1.5} model. The best score for each metric is highlighted in bold.
    }
    \label{tab:results3}
\end{table}

\stitle{Evaluating Training Strategy} As shown in Table~\ref{tab:results1}, we evaluated Med-VRAgent on medical VQA tasks using the LLaVA-Med v1.5 model, comparing it with five baselines: Zero-shot, SFT, Self-Rewarding, DPO, and STLLaVA-Med. On VQA-RAD, Med-VRAgent achieved 35.70 (open) and 68.72 (closed); on MIMIC-CXR, it scored 3.90 (BLEU), 13.53 (ROUGE-L), and 9.58 (METEOR), outperforming other methods in generalization and generation quality.

\stitle{Evaluating Reasoning Strategy} As shown in Table~\ref{tab:results2}, we tested Med-VRAgent's reasoning strategy, hypothesizing that improved visual guidance and feedback and higher-quality auxiliary information enhance performance. On the DeepSeek-VL-7B and MiniCPM-V2 models, Med-VRAgent outperformed others, achieving top scores in BVR (51.82 and 47.32) and average (46.74 and 43.41). Compared to Zero-shot, CoT, and ToT, it excelled in abnormality recognition, visual reasoning, and relational understanding, confirming the effectiveness of the Med-VRAgent in complex medical VQA.

\stitle{Performance Comparison of Decoding-based and RAG-based Methods} As shown in Table~\ref{tab:results3}, on the IU-Xray dataset, LLaVA-Med v1.5 performed poorly (BLEU=9.64), with modest improvements from Greedy and Beam Search (BLEU=12.10). MMed-RAG showed significant improvement (BLEU=31.38), while Med-VRAgent achieved the best results (BLEU=33.45, ROUGE-L=26.81, METEOR=33.12), demonstrating that Med-VRAgent enhances medical report generation quality.

\section{Discussion}

This section presents three experiments examining Med-VRAgent's performance in medical visual reasoning tasks. The first investigates the importance of each component. The second explores the impact of MCTS width and depth on model accuracy. The third experiment evaluates the adaptive retrieval strategy (ARS) in the Reflection component compared with the traditional fixed Top-K method.

\subsection{Analysis of Med-VRAgent's Components}

\begin{figure}
    \centering
    \begin{adjustbox}{width=1\linewidth}
        \includegraphics[width=\textwidth]{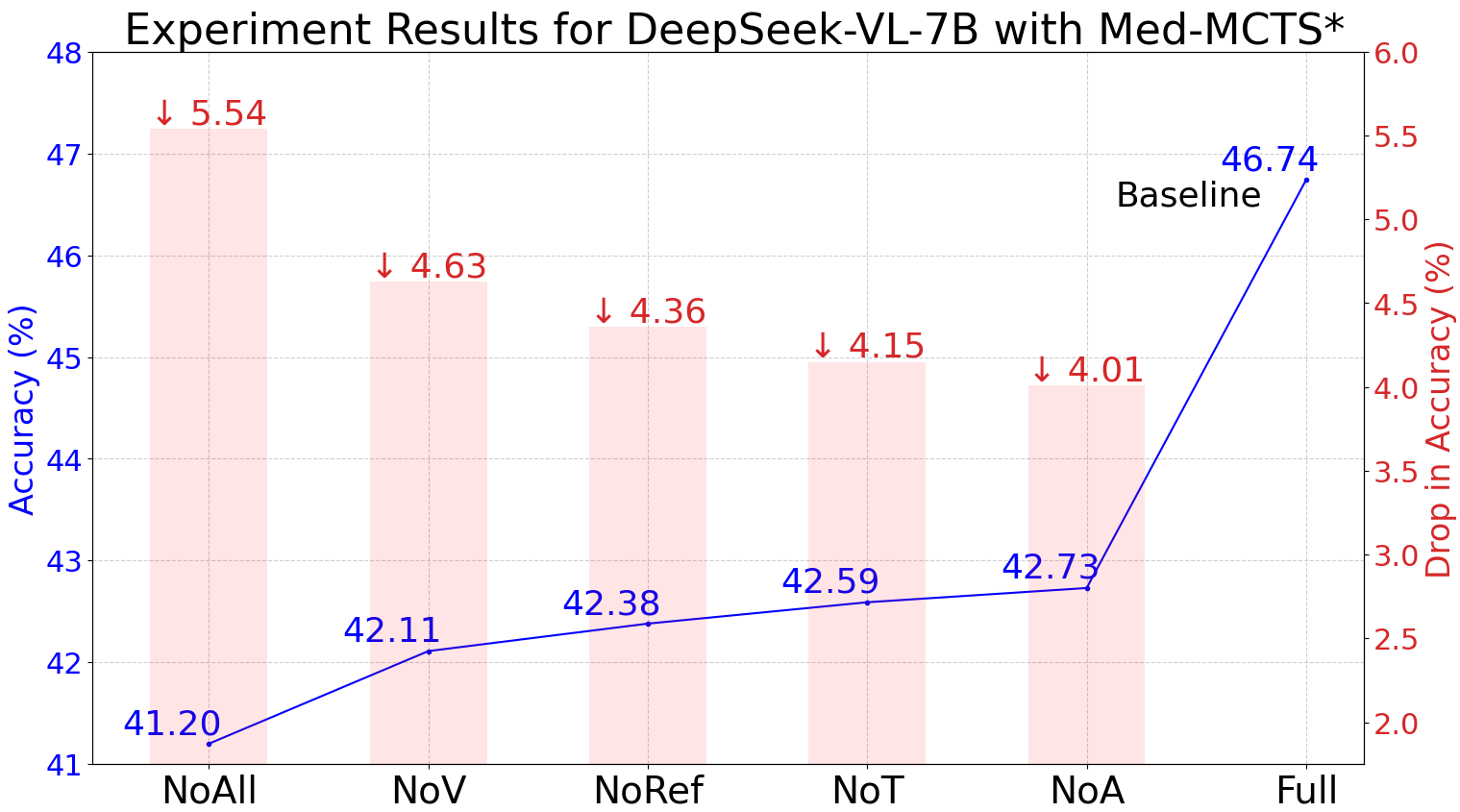}
    \end{adjustbox}
    \caption{Ablation Experiment 1 Results (accuracy; \%) for DeepSeek-VL-7B with Med-VRAgent on dataset GMAI-MMBench. Noall means removing all components, NoV means removing visual extraction, NOA means removing Assessor, and NoT means removing Teacher.}
    \label{fig:Ablation 1}
\end{figure}

We conduct an ablation study on \textbf{Med-VRAgent} to assess the contribution of its key components to medical visual reasoning. As shown in Fig~\ref{fig:Ablation 1}, removing any component leads to performance degradation, highlighting the critical role of each module in reasoning progression, relevance, coherence, and adaptability. The visual extraction component has the greatest impact. Specifically, omitting any module increases the error rate in LLMs, affecting reasoning quality.

\subsection{Width and Depth Optimization}

\begin{figure}
    \centering
    \begin{adjustbox}{width=1\linewidth}
        \includegraphics[width=\textwidth]{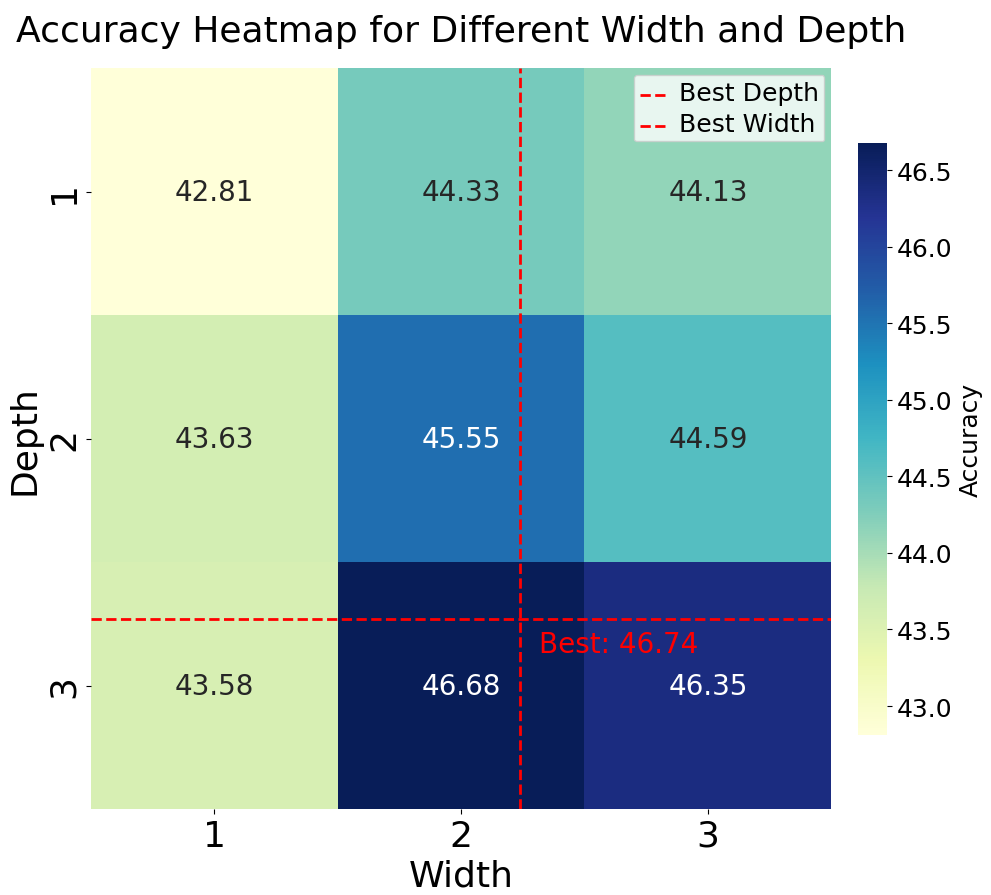}
    \end{adjustbox}
    \caption{ Ablation Experiment 2 Results (accuracy; \%)  for DeepSeek-VL-7B with Med-VRAgent on dataset GMAI-MMBench. Best is an adaptive exploration strategy, the average width and depth are 1.74 and 2.23 (red line), and other combinations are fixed width and depth. }
    \label{fig:Ablation 2}
\end{figure}

\begin{figure*}[htbp]
    \centering
    \begin{adjustbox}{width=\linewidth}
        \includegraphics[width=\textwidth]{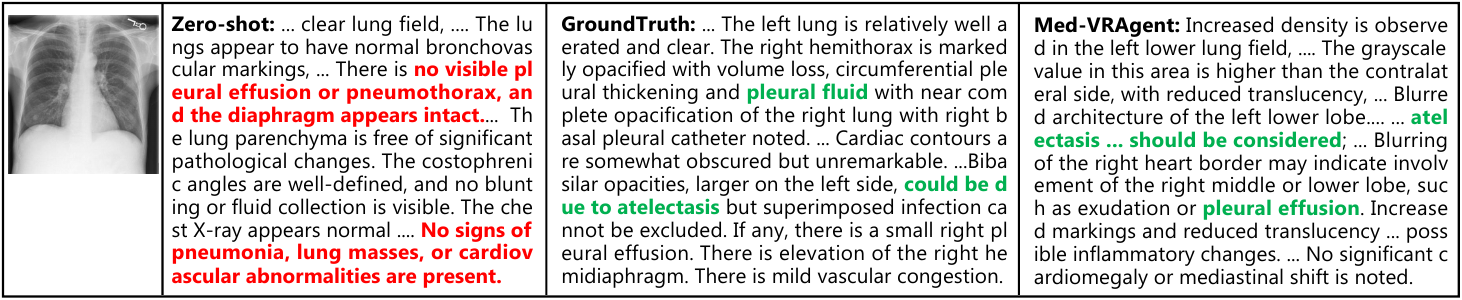}
    \end{adjustbox}
    \caption{ Med-VRAgent Medical Report Generation Case Study }
    \label{fig:Case}
\end{figure*}
We studied the impact of different MCTS fixed widths and depths on performance. The results are shown Fig~\ref{fig:Ablation 2}. By adjusting the fixed width and depth in the search strategy, we found that accuracy could be improved. Search benefits decrease as width and depth rise, likely due to VLM's limited processing capacity. The best fixed combination (width 2, depth 3) achieved the highest accuracy of 46.68\%. The adaptive strategy (width 1.74, depth 2.23) achieved an even higher accuracy of 46.74\%. This result demonstrates that our adaptive strategy can maintain a balance between exploration and exploitation.

\subsection{Evaluation of Adaptive Retrieval Strategy}
The Table~\ref{fig:Ablation 3} presents an ablation study on the MIMIC-CXR dataset using the LLaVA-Med v1.5 model, evaluating different retrieval strategies. The experiments compare the impact of enabling filtering and Rerank mechanisms on the quality of generated outputs. The results indicate that using either Filter or Rerank alone leads to modest performance improvements. For instance, compared to the Fixed Top-K baseline, the Rerank Only strategy shows slight gains across all metrics .  The best performance is Adaptive Retrieval, which combines both Filter and Rerank. It obtains the highest scores across all metrics. 

\begin{table}
    \centering
    \begin{adjustbox}{width=\linewidth}
        \setlength{\tabcolsep}{1pt}
        \tiny
\begin{tabular}{lccccc}
    \toprule
    \textbf{Experiment}  & \textbf{Filter} & \textbf{Rerank} & \textbf{BLEU} $\uparrow$ & \textbf{ROUGE-L} $\uparrow$ & \textbf{METEOR} $\uparrow$ \\
    \midrule
    Fixed Top-K         & \ding{55} & \ding{55} & 13.66  & 13.10  & 8.94  \\
    Rerank Only       & \ding{55} & \ding{51} & 13.75  & 13.20  & 9.22   \\
    Dynamic Top-K      & \ding{51} & \ding{55} & 13.80  & 13.75  & 9.12  \\
    Adaptive Retrieval  & \ding{51} & \ding{51} & \textbf{13.90}  & \textbf{14.10}  & \textbf{9.58} \\
    \bottomrule
\end{tabular}
    \end{adjustbox}
    \caption{
Ablation study on the MIMIC-CXR dataset using the LLaVA-Med v1.5 model. Each retrieval strategy varies in its use of Filter and Rerank.}
    \label{fig:Ablation 3}
\end{table}

\subsection{Performance and Efficiency Analysis}
In this experiment, we compared the performance of four methods (CoT, ToT, Med-VRAgent (Fix), Med-VRAgent (Ours)) on the GMAI-MMBench dataset. Fix is a fixed width of 2 and depth of 3. The results show that Med-VRAgent (Ours) performs best in terms of accuracy, reaching 46.74\%. In addition, Med-VRAgent (Ours) has an advantage over Med-VRAgent (Fix) in inference time, which is 36.7 seconds, significantly lower than the fixed strategy of 45.7 seconds. Although the ToT method is slightly higher than CoT in accuracy (42.08\% vs. 41.52\%), its inference time is longer, reaching 31.3 seconds. The Cot method is the most efficient in inference time, only 18.3 seconds, but its accuracy is lower. Overall, Med-VRAgent (Ours) has achieved a good balance between accuracy and inference time, showing its comprehensive advantages over fixed strategies and other methods. This shows that adaptive strategies can optimize inference time while improving accuracy, and have better application potential.

\begin{table}
\centering
  \begin{adjustbox}{width=\linewidth}
        \setlength{\tabcolsep}{1pt}
        \tiny
\setlength{\tabcolsep}{1pt}
\begin{tabular}{lcc}
\toprule
\textbf{Method} & \textbf{Accuracy (\%)} & \textbf{Inference Time (s)}   \\
\midrule
CoT  & 41.52 & \textbf{18.3} \\
ToT  & 42.08 & 31.3  \\
Med-VRAgent (Fix) & 46.68 & 45.7  \\
Med-VRAgent (Ours) & \textbf{46.74} & 36.7  \\
\bottomrule
\end{tabular}
\end{adjustbox}
\caption{DeepSeek-VL-7B compares the inference accuracy and average time of  CoT, ToT and Med-VRAgent (fixed and adaptive policies)  on the GMAI-MMBench dataset. }
\label{tab:efficiency_comparison}
\end{table}

\subsection{Case Study}
As shown in the Fig~\ref{fig:Case}, the case comes from the Deepseek-VL and the MIMI-CXR dataset. Med-VRAgent outperforms the Zero-shot in generating clinically accurate and factually grounded chest X-ray reports. While the Zero-shot model incorrectly states clear lungs and no pleural abnormalities, Med-VRAgent correctly identifies increased density, reduced translucency, and possible pleural effusion in the left lung, closely matching the expert GroundTruth. It avoids major hallucinations and captures subtle findings like blurred architecture and right heart border changes, suggesting infection or inflammation. Med-VRAgent also includes diagnostic considerations such as atelectasis, reflecting expert-level reasoning. 

\section{Conclusion}

This study introduces Med-VRAgent, a novel medical visual reasoning framework that enhances multimodal large models' performance in medical image understanding. It incorporates a teacher-student-evaluator mechanism, visual guidance and self-feedback paradigm, and a multi-step reasoning strategy based on MCTS. Med-VRAgent achieved top performance across several medical multimodal benchmark datasets, demonstrating proficiency in image-text alignment, spatial structure understanding, and lesion recognition. Future research will focus on improving search efficiency, using advanced multimodal models, and expanding deployment in real clinical settings.

\section*{Limitations}
Although Med-VRAgent has achieved significant improvements in medical visual reasoning, it still has limitations. Despite optimization, tree search is still resource-intensive. Due to node expansion strategies and computational resource constraints, it may not be possible to fully search all possible reasoning paths. It may not be directly transferable to other domains and additional domain adaptation is required. Visual guidance may have limited effect in complex images or low-quality images. Inaccurate reasoning may still occur when faced with fine-grained errors or very complex cases. Performance and reliability in actual clinical settings have not been fully verified.

\section*{Ethical Considerations}
Ethical considerations are central to our research. In this study, we ensure adherence to ethical principles by exclusively using publicly available datasets and employing models that are open-source or widely accepted within the research community. We emphasize transparency in all stages of our work and prioritize the responsible application of technology, particularly in the sensitive domain of medical reasoning, to ensure that our contributions promote fairness, reliability, and societal benefit.

\bibliography{acl2020}
\bibliographystyle{acl_natbib}

\newpage
\appendix
\onecolumn

\label{sec:Comparedmethods}

\section{ Prompt } 

\begin{figure*}[h]
    \begin{userquery}
    \fontsize{10}{10}\selectfont
Role Setting:
You are a medical expert providing guidance on medical image analysis to help students improve their understanding.

Task Description:
Focus on the red-boxed area in the image, using previous guidance and student feedback to offer optimized suggestions for enhancing their analysis skills.

Guidance Content:

Analyze Key Area:

Identify the red-boxed region for closer analysis.

Observe structural features, shape changes, color contrasts, and any abnormalities.

Reference Feedback and Suggestions:

Evaluate the student's previous analysis.

Point out missed details or inadequate analysis, and offer visual techniques.

Optimize Analysis Directions:

Guide the student based on the image type (e.g., CT, X-ray, ultrasound).

Suggest perspectives like cross-sections or tissue density changes.

Important Notes:

Your goal is to help students master image analysis, not to do it for them.

Focus on a logical, systematic approach for comprehensive image interpretation.

Previous Guidance:
</Guidance>

Student’s Answer:
</Answer>

Feedback Information:
</Feedback >

Use this format for guidance:
</Guidance> Guidance here </Guidance>

\end{userquery}
\caption{ROI-Guided Teaching Prompt}
\label{fig: system_prompt_1}
\end{figure*}

\begin{figure*}[h]
    \begin{userquery}
    \fontsize{10}{10}\selectfont
You are a medical expert. Please review the image and visual analysis guidance and rate the student-generated answers using the additional 5-point rating system described below. The rating will be cumulative based on the following criteria:

5-point rating scale:

1. Relevant information: If the medical vision answer provides some information that is relevant to the user's query, even if the information is incomplete or contains some incompletely relevant content, 1 point can be awarded.
2. Partially solve the problem: If the answer solves most of the user's question, but does not fully answer the user's question or does not directly answer the core query, 2 points can be awarded.
3. Essential elements: If the answer answers the basic elements of the user's question from a medical vision perspective, although it may lack detail or completeness in some aspects, but is still helpful to the user, 3 points can be awarded.
4. Direct and comprehensive solution to the problem: If the answer directly and comprehensively solves the user's question, although there may be some room for improvement in clarity, conciseness or visual focus, 4 points can be awarded.
5. Tailored, professional and profound: If the answer is tailored to the user's question, provides an in-depth and professional answer through medical vision, avoids irrelevant information, and produces high-quality, engaging and insightful content, 5 points should be awarded.

Information: <guidance> {Teacher's guidance} </guidance> <answer> {Student's answer} </answer>

Evaluation steps:

Total rating: Please briefly explain your rating in 100 words or less. 

Suggestions for teachers: Provide suggestions for teachers to build better guidance in 100 words or less.

Revision suggestions for students: Provide revision suggestions for students in 100 words or less.

Rating conclusion:

<score>{Integer score}</score>

<feedback1>{Feedback to teachers}</feedback1>

<feedback2>{Revision suggestions for students}</feedback2>

\end{userquery}
\caption{ROI-Guided Evaluation Prompt}
\label{fig: system_prompt_2}
\end{figure*}

\newpage
\section{Med-VRAgent algorithm }  
\begin{algorithm}[H]
\small 

\SetAlCapFnt{\normalsize}
\SetAlCapNameFnt{\normalsize}
\caption{Med-VRAgent}

\KwIn{Question $Q$, Image $I$, Visual extractor $\mathcal{V}$, Teacher $T$, Assessor $A$, Student $S$, Retriever $R$, max\_depth $D_{\mathrm{epth}}$, max\_branch\_number $b$, max\_simulation\_number $Sim$}

\KwOut{Best solution path $\pi^\star$; Final answer $A_{\mathrm{final}}$}

$\mathbf{ROI} \gets \mathcal{V}(I,Q)$ \tcp*{Region-of-interest detection}

$\mathcal{T} \gets \textsc{InitializeTree}(Q, I)$

\For{$t=1$ \KwTo $Sim$}{
    $C \gets \textsc{Root}(\mathcal{T})$\;

    \tcc{--- Selection ---}
    \While(\tcp*[f]{C is not a leaf}){C is not a leaf node}{
        $C \gets \arg\max_{s}\mathrm{UCB}(s)$\;
        \If(\tcp*[f]{max depth reached}){$\text{depth}(C) \ge D_{\mathrm{epth}}$}{
            \textbf{break}\;
        }
        \If(\tcp*[f]{node not fully expanded}){$C$ has less than $b$ children nodes}{
            \textbf{break}\;
        }
    }

    \If(\tcp*[f]{skip if depth limit}){$\text{depth}(C) \ge D_{\mathrm{epth}}$}{
        \textbf{continue}\;
    }

    \tcc{--- Expansion \& Evaluation ---}
    $O^{\text{anc}}_{g} \gets \bigcup_{k\in\mathrm{ancestor}(C)} G_k$ \tcp*{Teacher’s guidance from ancestor}
    $O^{\text{anc}}_{a} \gets \bigcup_{k\in\mathrm{ancestor}(C)} A_k$ \tcp*{Student’s answers from ancestor}
    $O^{\text{sib}}_{g} \gets \bigcup_{k\in\mathrm{siblings}(C)} G_k$ \tcp*{Teacher’s guidance from siblings}
    $O^{\text{sib}}_{a} \gets \bigcup_{k\in\mathrm{siblings}(C)} A_k$ \tcp*{Student’s answers from siblings}
    $O^{\text{sib}}_{f} \gets \bigcup_{k\in\mathrm{siblings}(C)} F_k$ \tcp*{Assessor’s feedback from siblings}

    $O \gets (O^{\text{anc}}_{g}, O^{\text{anc}}_{a}, O^{\text{sib}}_{g}, O^{\text{sib}}_{a}, O^{\text{sib}}_{f})$\;

    $roi \gets \textsc{SelectOnProb}(\textsc{P\_softmax}(Conf_{\text{roi}}))$\;
    $G \gets T(\textit{roi}, O)$ \tcp*{Generate guidance (§3.3)}
    $A \gets S(\textit{roi}, G)$ \tcp*{Student answer (§3.3)}
    $(R,F) \gets A(\textit{roi}, G, A)$ \tcp*{Score \& feedback (§3.3)}

    \If(\tcp*[f]{Stop early if full 5-point score}){$R == 5$}{
        \textbf{break}\;
    }

    $C' \gets \textsc{CreateNewChild}(G, A, R, F, O)$ \tcp*{Create new child node for $C$}
    \textsc{AddChild}(C,C') \tcp*{Add $C'$ to the children of $C$}\;

    \tcc{--- Backpropagation ---}
    \textsc{Backpropagate}$(C)$ \tcp*{Update visit-count \& reward}
}

$\pi^\star \gets \textsc{BestPath}(\mathcal{T})$ \tcp*{Highest cumulative reward}

\tcc{--- Reflection ---}
\For{node in $\pi^\star$}{
    \If{$R < 4$}{
        $\mathcal{K} \gets \textsc{Rerank}(\textsc{Relevance}(\textsc{TopK}(A, G, I)))$ \tcp*{Retrieval (§3.4)}
        $A^\star \gets S(\textit{roi}, G, A, \mathcal{K})$ \tcp*{Rewrite}
        \textsc{UpdateNode}$(A^\star)$
    }
}

$A_{\mathrm{final}} \gets \textsc{ComposeAnswer}(\pi^\star)$\;
\KwRet{$\pi^\star, A_{\mathrm{final}}$}\;

\end{algorithm}

\onecolumn

\section{Ablation Studies}

\subsection{Visual Token Edit Ablation Results}

\begin{table}[htbp]
\centering

\begin{tabular}{lccc}
\toprule
Method & BOX & Edit & Accuracy \\
\midrule
No VTE           & no  & no  & 42.11 \\
Only BOX No Edit & yes & no  & 43.22 \\
Only Edit No BOX & no  & yes & 45.56 \\
VTE              & yes & yes & 46.74 \\
\bottomrule
\end{tabular}
\caption{Visual Token Edit Ablation Results for DeepSeek-VL-7B (Student) with Med-VRAgent on GMAI-MMBench. }
\end{table}

BOX refers to bounding box prompts on ROI, and Edit refers to attention enhancement on ROI. The results show that using both simultaneously yields the best performance, while omitting either leads to performance drops.

\subsection{Teacher Guidance Ablation Results}

\begin{table}[htbp]
\centering

\begin{tabular}{lcccc}
\toprule
Method & Guidance & Answer & Feedback & Accuracy \\
\midrule
1 & no  & no  & no  & 42.03 \\
2 & yes & no  & no  & 42.31 \\
3 & yes & yes & no  & 43.01 \\
4 & yes & no  & yes & 42.64 \\
5 & yes & yes & yes & 43.41 \\
\bottomrule
\end{tabular}
\caption{Teacher Guidance Ablation Results for MiniCPM-V2 with Med-VRAgent on GMAI-MMBench.}
\end{table}

The three middle columns denote the information available to the teacher. For example, Method 1 indicates the teacher sees nothing and merely samples multiple times. Rows 2--5 progressively allow the teacher to access prior guidance, the student’s answers, and feedback from the assessor, validating the effectiveness of heuristic-based teacher guidance.

\subsection{GREEN Evaluation Results}

\begin{table}[htbp]
\centering
\begin{tabular}{lc}
\toprule
Method & GREEN \\
\midrule
Zero-Shot          & 0.21 \\
RULE               & 0.29 \\
MMed-RAG           & 0.31 \\
Med-VRAgent (ours) & 0.34 \\
\bottomrule
\end{tabular}
\caption{GREEN scores of LLaVA-Medv1.5 on IU-Xray.}
\end{table}

We performed a preliminary evaluation using the GREEN~\cite{Ostmeier_2024} approach. GREEN uses LLMs as evaluators and provides scores that are consistent with expert preferences and human-interpretable for clinically significant errors (0 to 1, higher is better). We sampled 100 examples from the IU-Xray dataset. We will add more new experimental results later.

\end{document}